\documentclass[journal]{IEEEtran}
\usepackage{amsmath,amsfonts,amssymb}
\usepackage{algorithmic}
\usepackage{algorithm}
\usepackage{array}
\usepackage[caption=false,font=normalsize,labelfont=sf,textfont=sf]{subfig}
\usepackage{textcomp}
\usepackage{stfloats}
\usepackage{url}
\usepackage{verbatim}
\usepackage{graphicx}
\usepackage{cite}
\usepackage{booktabs}
\usepackage{multirow}
\usepackage{xcolor}
\usepackage{enumitem}
\usepackage{placeins}

\begin{document}
\bstctlcite{IEEEtranBSTCTL:etal}

\title{DeicticVLA: Unifying Instruction Modes Based on \\ Language and Deictic Gestures in a Single VLA}

\author{Kango Yanagida$^{1}$ Tatsuya Aoki$^{1}$ Yuichiro Yoshikawa$^{1}$ and Takato Horii$^{1,2}$  
\thanks{*This work was supported by JST Moonshot R\&D Program Japan Grant Number JPMJMS2011, and by JST BOOST, Japan Grant Number JPMJBS2402.}%
\thanks{$^{1}$Dept. of Systems Innovation, Graduate School of Engineering Science, The University of Osaka, Japan.}%
\thanks{$^{2}$International Research Center for Neurointelligence, The University of Tokyo, Japan.}%
\thanks{\tt\small \{yanagida.kango.z7m@ecs., aoki.tatsuya.es@, y.yoshikawa.es@, takato@sys.es.\} osaka-u.ac.jp}
}%

\maketitle

\begin{abstract}
Vision-Language-Action models (VLAs) allow users to specify manipulation tasks in natural language, but distinguishing a target or placement goal among objects of the same category or similar appearance requires detailed expressions that VLAs may not use reliably. We propose DeicticVLA, which canonicalizes Language Instruction (LI), Vision-Language Instruction (VLI), and Visual Instruction (VI) into a text prompt and deictic masks through text-prompt completion and deictic gesture grounding, enabling a single pretrained VLA to handle all three instruction modes. With a shared backbone, demonstrations, and matched training steps, we compare two RGB visual prompting methods, two separate-channel mask prompting methods, and three training strategies in simulation. Under two-stage training, the four prompting methods achieve high in-distribution success but differ in their ability to use deictic masks in unseen layouts. Across methods, training-strategy ablations show that two-stage training improves such use, while retaining second-stage LI data mitigates forgetting without reducing VLI and VI performance. In three real-world tasks, one policy supports all modes. VLI and VI outperform LI under unseen expressions, appearance changes, and novel objects. For unseen categories, both achieve 100\% success, compared with 16.7\% for jointly trained LI. These results demonstrate the unified three-mode interface and guide DeicticVLA design.
\end{abstract}


\section{Introduction}

\IEEEPARstart{V}{ision}-Language-Action models (VLAs) provide a promising framework for transferring the rich knowledge of pretrained Vision-Language Models to robot control, thereby enabling users to specify manipulation tasks through natural language~\cite{rt2, pi0}. However, real-world environments such as homes and offices often contain multiple objects of the same category or with similar appearances. Uniquely identifying a target using language alone therefore requires detailed referring expressions involving appearance, spatial relations, or ordinal terms, increasing the descriptive burden on the user. At the same time, VLAs do not necessarily use such detailed linguistic expressions reliably and have been reported to select manipulation targets based more strongly on visual biases in the training data than on language tokens~\cite{langgap, counterfactual}. They may also fail to ground spatial relations and ordinal expressions~\cite{libero-plus}. Language-only task specification thus poses a double problem: it requires users to provide detailed descriptions to identify a target uniquely, yet the model may not use those descriptions correctly.

\begin{figure*}[t]
    \centering
    \includegraphics[width=\linewidth]{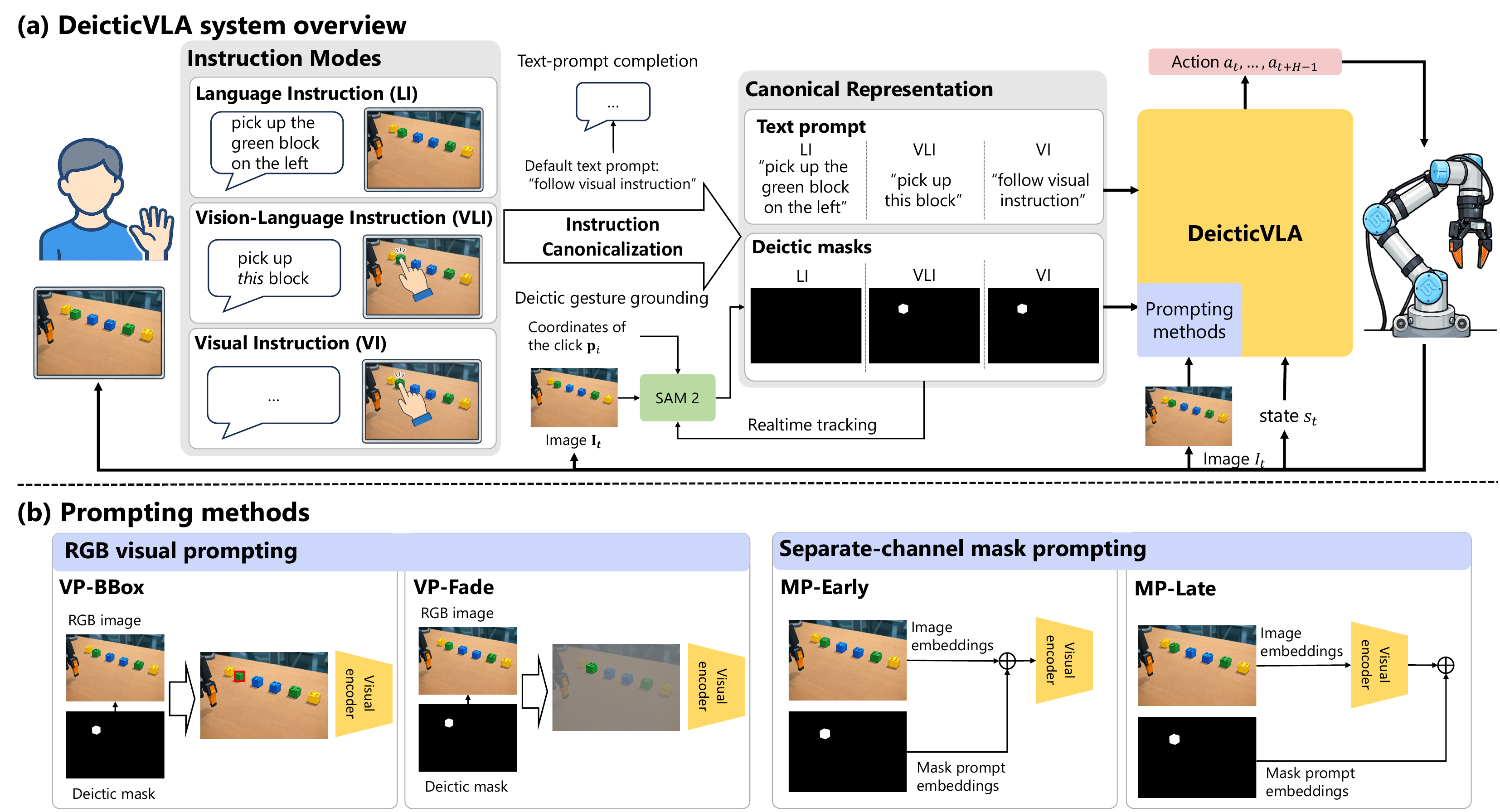}
    \caption{Overview of the DeicticVLA system and the prompting methods evaluated in this work. (a) The user specifies a task through LI, VLI, or VI, which provide language only, language with a deictic gesture, or a deictic gesture only, respectively. Instruction canonicalization converts these user inputs into a canonical representation consisting of a text prompt and deictic masks through text-prompt completion and deictic gesture grounding. DeicticVLA uses this representation together with RGB observations and the robot state to generate an action chunk. (b) VP-BBox and VP-Fade construct RGB visual prompts by drawing a bounding box or fading regions outside the deictic mask on the RGB observation, respectively. MP-Early and MP-Late encode the mask separately and fuse the resulting mask prompt embeddings before and after the visual encoder, respectively.}
    \label{fig:system}
\end{figure*}

In human communication, demonstratives such as ``this'' and ``there'' are combined with deictic gestures such as pointing to directly indicate a target in a shared perceptual space and resolve references that would be ambiguous in language alone. Introducing this mechanism into VLAs is a natural approach to the double problem described above. However, two challenges remain.

First, no established method handles within a single policy the three instruction modes of Language Instruction (LI), in which the user provides language alone; Vision-Language Instruction (VLI), in which language is combined with a deictic gesture; and Visual Instruction (VI), in which only a deictic gesture is provided. Most previous studies applying deictic gestures to robot manipulation are limited to either VLI or VI~\cite{moo, this_and_that, sketch_moma, rovi, gaze2act, gesvla, vca}. Although Point-VLA~\cite{pointvla} supports LI and VLI, no single unified policy has been proposed that allows switching among all three instruction modes, including VI.

Second, no established design principle determines how region information obtained from a deictic gesture should be presented to and learned by a pretrained VLA. Existing studies have used RGB visual prompting, which renders the target region on the RGB observation~\cite{Hannus2025IAVLA, bringmycup, pointvla, gesvla}, and separate-channel mask prompting, which receives the mask separately from RGB and fuses it at the feature level~\cite{moo, gaze2act, controlvla}. However, because these studies differ in their backbones, data composition, and training strategies, the effect of the conditioning method itself cannot be isolated. To our knowledge, no study has systematically compared RGB visual prompting with separate-channel mask prompting at different fusion positions using the same pretrained VLA and demonstrations, with the total number of training steps matched, while also examining the training strategy.

To address these challenges, we propose DeicticVLA, a unified interface that canonicalizes user inputs provided through LI, VLI, and VI into a common representation consisting of a text prompt and deictic masks, allowing all three modes to be handled by a single pretrained VLA. Instruction canonicalization consists of text-prompt completion and deictic gesture grounding, in which user clicks in VLI and VI are grounded into deictic masks by SAM~2~\cite{sam2}. After canonicalization, the deictic masks are supplied to the VLA either as visual prompts rendered on the RGB observation or as mask prompts provided through a pathway separate from RGB. Within this controlled comparison, we evaluate two RGB visual prompting methods and two separate-channel mask prompting methods in combination with single-stage and two-stage training strategies. In two-stage training, the model is first adapted to the task using LI data and is then jointly trained on LI, VLI, and VI. We further evaluate the model on three real-world tasks and show that VLI and VI substantially outperform LI under instructions with unseen spatial references and ordinal expressions, under visual perturbations, and with unseen objects.

The contributions of this study are as follows:
\begin{enumerate}
\item We propose DeicticVLA, a unified interface that canonicalizes user inputs provided through LI, VLI, and VI into a common representation consisting of a text prompt and deictic masks, allowing switching among the three instruction modes within a single pretrained VLA.
\item We introduce two-stage training that first adapts the VLA to tasks using LI data and then jointly trains it on LI, VLI, and VI. Using the same backbone and demonstrations, with the total number of training steps matched, we systematically compare two RGB visual prompting methods and two separate-channel mask prompting methods, and ablate the number of training stages and the inclusion of LI data in the second stage to clarify their effects on task success and generalization to unseen conditions.
\item We validate DeicticVLA on three real-world manipulation tasks and show that VLI and VI achieve higher task success rates than LI for unseen instruction expressions and unseen objects.
\end{enumerate}

\section{Related Work}
\subsection{Deictic Gesture Interfaces for Object Specification}
Interfaces that allow users to specify objects or regions involved in manipulation tasks to a robot through nonverbal deictic actions have been explored across various modalities. Examples include sketches~\cite{sketch_moma, rovi, rt_trajectory}, clicks~\cite{this_and_that, tognet, vca}, click-and-drag bounding boxes~\cite{bbox_act, pointvla}, pointing gestures~\cite{gesvla}, and gaze~\cite{gaze2act}, all of which directly indicate objects or regions in the robot's current field of view. In contrast, presenting a reference image~\cite{bringmycup, interleavevla} or a goal image~\cite{oevla} visually describes a target's appearance or goal state and is therefore distinct from deixis, which indicates a region in the current view. Among these deictic modalities, we adopt clicking. A click specifies a single coordinate on a device displaying the robot-view image, entails less inter-user variation in input shape than a sketch, and can be converted into a pixel-level region by an existing segmentation model. In this study, a deictic mask generated by SAM~2~\cite{sam2} from click coordinates serves as the common source of visual prompts and mask prompts.

\subsection{Multimodal Task Specification for Robot Manipulation}
Most previous studies fix the instruction mode through which a user specifies a task to a robot policy. The first is LI, which includes conventional VLAs~\cite{rt2, pi0}. Methods in which a VLM or another module automatically generates visual guidance from a text prompt and supplies it to a policy~\cite{Hannus2025IAVLA, byovla, roboground, controlvla}, as well as methods in which the model itself outputs regions during reasoning~\cite{ecot, gemini_robotics}, also fall under LI from the user's perspective. In these methods, the user's input remains language, so the burden of composing a description that uniquely identifies a target when multiple objects of the same category or similar appearance are present remains. The second is VLI, in which the user combines language with a deictic gesture so that nonverbal target specification complements linguistic ambiguity~\cite{this_and_that, gaze2act, gesvla}. The third is VI, in which the user specifies a task through a deictic gesture without providing task-specific language~\cite{vca, bbox_act}.
In contrast, few studies support multiple instruction modes within a single model. Point-VLA~\cite{pointvla} supports two instruction modes corresponding to LI and VLI but has no mode corresponding to VI. To our knowledge, no previous study has unified within a single pretrained VLA the three instruction modes of LI using language alone, VLI combining language with a deictic gesture that indicates a region in the current view, and VI using only that deictic gesture.

\subsection{Visual and Mask Prompting for Manipulation Policies}
Visually conditioning a model on task-relevant objects or regions has a long history. In VLMs, visual prompting that overlays visual markers on RGB images is widely used~\cite{redcircle, vipllava, som}. Manipulation policies have likewise used methods that render trajectories~\cite{rt_trajectory, tracevla}, anchors~\cite{vpvla}, bounding boxes~\cite{bbox_act, pointvla}, or segmentation masks~\cite{bringmycup, Hannus2025IAVLA} on RGB images. Even when using the same type of mask, BringMyCup~\cite{bringmycup} overlays color on the target object, whereas IA-VLA~\cite{Hannus2025IAVLA} grays out the background outside the target.
Another approach is separate-channel mask prompting, which receives a mask as an input separate from RGB, namely as a mask prompt, and fuses it at the feature level. Alpha-CLIP~\cite{alphaclip} and Describe Anything~\cite{dam} adopt this pathway in VLMs. In manipulation policies, it has long been used in policies trained from scratch with object-centric representations~\cite{groot, tpm, tognet, roboground, vca} and has recently been extended to pretrained VLAs~\cite{moo, controlvla, gaze2act}.

To our knowledge, no previous study has systematically compared RGB visual prompting with separate-channel mask prompting at different fusion positions using the same pretrained VLA and demonstrations, with the total number of training steps matched. We implement two methods that generate visual prompts from deictic masks and render them on RGB images, as well as two methods that fuse deictic masks as mask prompts before or after the visual encoder, and compare them in combination with training strategies in Section~\ref{sec:sim}. An exploratory configuration that repeatedly fuses the mask prompt at intermediate layers is reported in Appendix~\ref{app:mp_inter}.


\section{Proposed Method}

Fig.~\ref{fig:system}(a) presents the system overview of DeicticVLA. The user specifies a task through one of three instruction modes, LI, VLI, or VI, which differ in whether language and a deictic gesture are provided. Instruction canonicalization, consisting of text-prompt completion and deictic gesture grounding, transforms these user inputs into a canonical representation comprising a text prompt and deictic masks. This representation is then provided to DeicticVLA. Depending on the prompting method shown in Fig.~\ref{fig:system}(b), DeicticVLA uses the deictic masks either as visual prompts rendered on the RGB observations or as mask prompts supplied through a pathway separate from RGB.

\subsection{Preliminaries}
A standard VLA policy generates an action chunk of length $H$ conditioned on the RGB observation $\mathbf I_t$, robot state $\mathbf s_t$, and text prompt $\ell_t$ at time $t$. Here, $\mathbf I_t$ denotes RGB images obtained from one or more cameras. A VLA policy $\pi_\theta$ with parameters $\theta$ is expressed as
\begin{equation}
\hat{\mathbf A}_t
=
[\hat{\mathbf a}_t,\ldots,\hat{\mathbf a}_{t+H-1}]
\sim
\pi_\theta\!\left(
\cdot
\mid
\mathbf I_t,\mathbf s_t,\ell_t
\right).
\end{equation}

DeicticVLA adds the deictic masks $\mathbb M_t$, defined below, as a conditioning input without changing the output action space or the training objective of the base VLA.
\begin{equation}
\hat{\mathbf A}_t
\sim
\pi_\theta\!\left(
\cdot
\mid
\mathbf I_t,\mathbf s_t,\ell_t,\mathbb M_t
\right).
\end{equation}

\subsection{Instruction Modes and Canonicalization}

\subsubsection{User-Facing Instruction Modes}
DeicticVLA supports three instruction modes, LI, VLI, and VI. For mode $m$, let $\ell_{\mathrm{usr}}^{(m)}$ denote the language input provided by the user and $\mathbb P_{\mathrm{usr}}^{(m)}$ denote the set of click coordinates provided as a deictic gesture. We define the user input as
\begin{equation}
\left(\ell_{\mathrm{usr}}^{(m)},\mathbb P_{\mathrm{usr}}^{(m)}\right)
=
\begin{cases}
\left(\ell_{\mathrm{LI}},\varnothing\right), & m=\mathrm{LI},\\
\left(\ell_{\mathrm{VLI}},\mathbb P_{\mathrm{usr}}\right), & m=\mathrm{VLI},\\
\left(\varnothing,\mathbb P_{\mathrm{usr}}\right), & m=\mathrm{VI}.
\end{cases}
\end{equation}

\noindent Here, $\varnothing$ indicates that the user does not provide the corresponding input. In LI, the user provides only language $\ell_{\mathrm{LI}}$. In VLI, the user combines language $\ell_{\mathrm{VLI}}$ containing a deictic expression such as ``this'' or ``there'' with a deictic gesture. In VI, the user provides only a deictic gesture and no language. Although VI is limited to simple tasks such as pick-and-place, it enables concise specification of frequently performed routine manipulation tasks.

\subsubsection{Instruction Canonicalization}
Instruction canonicalization consists of text-prompt completion and deictic gesture grounding.

Text-prompt completion defines the text prompt provided to the VLA for every mode. LI and VLI use the language input supplied by the user without modification, whereas VI uses the default text prompt $\ell_0=\text{``follow visual instruction''}$. Thus,
\begin{equation}
\ell^{(m)}
=
\begin{cases}
\ell_{\mathrm{LI}}, & m=\mathrm{LI},\\
\ell_{\mathrm{VLI}}, & m=\mathrm{VLI},\\
\ell_0, & m=\mathrm{VI}.
\end{cases}
\end{equation}

\noindent The prompt $\ell_0$ is shared across all tasks and contains no task content such as the manipulated object, placement goal, or spatial relation. In VI, the deictic masks defined below specify the task.

In deictic gesture grounding, the user clicks task-relevant objects in the robot-view image $\mathbf I_t^{(\mathrm{ui})}$ displayed on a PC or tablet. Let $\mathbb P_{\mathrm{usr}}=\{\mathbf p_i\}_{i=1}^{n}$. We define the deictic mask obtained by grounding the $i$th click with SAM~2 as
\begin{equation}
\mathbf M_{t,i}
=
\mathcal S\left(\mathbf I_t^{(\mathrm{ui})},\mathbf p_i\right),
\quad
\mathbf M_{t,i}\in\{0,1\}^{H\times W},
\quad
i=1,\ldots,n.
\end{equation}

\begin{figure}[t]
    \centering
    \includegraphics[width=0.9\linewidth]{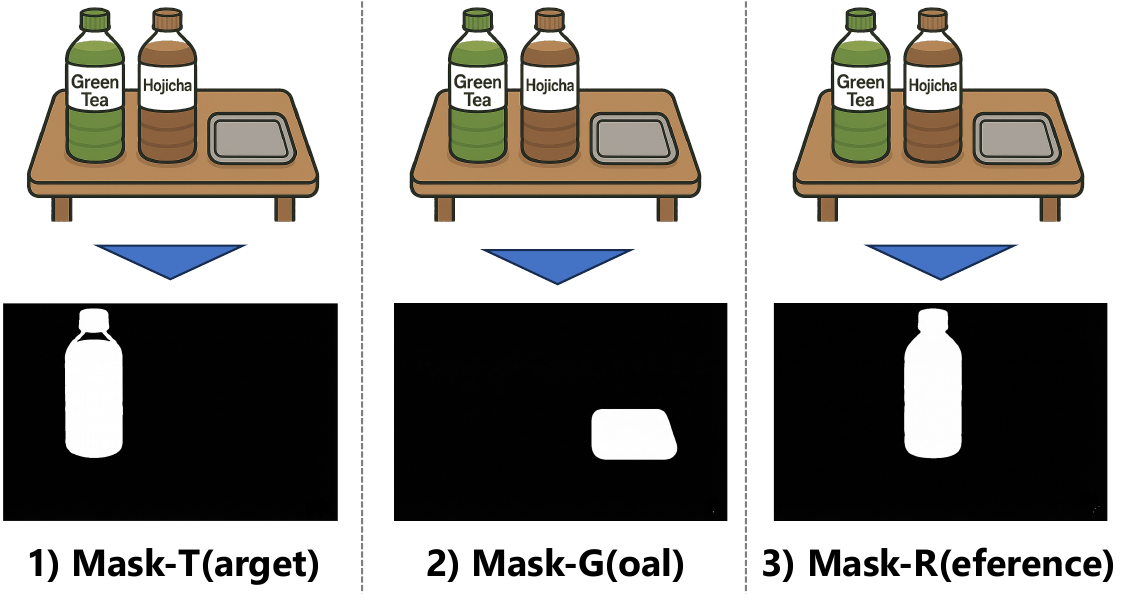}
    \caption{Roles of deictic masks in task specification. Mask-T specifies the object to be manipulated, Mask-G specifies the placement goal, and Mask-R specifies a reference object used in a spatial referring expression. For example, in ``put the green tea on the tray,'' the green-tea bottle is Mask-T and the tray is Mask-G. In ``put the tea next to this on the tray,'' the deictically indicated green-tea bottle serves as Mask-R.}
    \label{fig:mask_type}
\end{figure}

\noindent Here, $\mathcal S$ denotes segmentation and temporal tracking by SAM~2. We use \texttt{sam2.1-hiera-tiny}, which performs mask tracking at approximately 30~Hz on an NVIDIA RTX~4090. The generated deictic masks are retained separately as $\mathbb M_{t,\mathrm{usr}}=\left\{\mathbf M_{t,i}\right\}_{i=1}^{n}$. We denote the deictic masks as Mask-T, Mask-G, and Mask-R according to the role of the specified object in the task, as shown in Fig.~\ref{fig:mask_type}. \textbf{Mask-T(arget)} specifies the object to be grasped or moved. \textbf{Mask-G(oal)} specifies the region of the object on which the manipulated object is to be placed. \textbf{Mask-R(eference)} specifies an object that serves as the reference for a spatial referring expression. For example, Mask-T+G retains two masks, namely Mask-T for the manipulated object and Mask-G for the placement goal. Mask-R indicates the referent of ``this'' in an expression such as ``the tea next to this'' and is used primarily in VLI.

The outputs of text-prompt completion and deictic gesture grounding are standardized into a canonical representation consisting of a text prompt and deictic masks. We define the canonical representation for each mode as
\begin{equation}
\left(\ell^{(m)},\mathbb M_t^{(m)}\right)
=
\begin{cases}
\left(\ell_{\mathrm{LI}},\{\mathbf 0\}\right), & m=\mathrm{LI},\\
\left(\ell_{\mathrm{VLI}},\mathbb M_{t,\mathrm{usr}}\right), & m=\mathrm{VLI},\\
\left(\ell_0,\mathbb M_{t,\mathrm{usr}}\right), & m=\mathrm{VI}.
\end{cases}
\end{equation}

\noindent This canonical representation is a common upstream representation independent of the prompting method. The deictic masks contained in it are used as either visual prompts or mask prompts according to the prompting method described in the next subsection. We therefore distinguish among the user input, the canonical representation, and the specific prompts provided to the VLA.

\subsection{RGB Visual Prompting and Separate-Channel Mask Prompting}
\label{subsec:vla_and_robot_component}

DeicticVLA receives the canonical representation $\left(\ell^{(m)},\mathbb M_t^{(m)}\right)$ and incorporates the deictic masks into the visual input of the VLA according to the prompting method. All four prompting methods use the same canonical representation and thus the same text prompts and deictic masks; they differ only in how the masks are presented to the VLA. In the main text, we compare two RGB visual prompting methods, VP-BBox and VP-Fade, and two separate-channel mask prompting methods, MP-Early and MP-Late. The design of MP-Inter-$k$ and the results for MP-Inter-3 are presented in Appendix~\ref{app:mp_inter}.

In RGB visual prompting, rendering functions transform the RGB observations using the deictic masks. Let $\mathcal R_{\mathrm{Fade}}$ and $\mathcal R_{\mathrm{BBox}}$ denote the rendering functions of VP-Fade and VP-BBox, respectively. Their outputs are
\begin{equation}
\begin{aligned}
\widetilde{\mathbf I}_{t,\mathrm{Fade}}^{(m)}
&=
\mathcal R_{\mathrm{Fade}}\!\left(\mathbf I_t,\mathbb M_t^{(m)}\right),
\\
\widetilde{\mathbf I}_{t,\mathrm{BBox}}^{(m)}
&=
\mathcal R_{\mathrm{BBox}}\!\left(\mathbf I_t,\mathbb M_t^{(m)}\right).
\end{aligned}
\end{equation}

\noindent For LI, neither method renders a visual prompt, and $\widetilde{\mathbf I}_t^{(\mathrm{LI})}=\mathbf I_t$. \textbf{VP-Fade} follows IA-VLA~\cite{Hannus2025IAVLA} and fades all regions except the deictic masks and the robot region with semitransparent gray at an opacity of 0.8. \textbf{VP-BBox} draws the bounding box of each mask on the RGB observation. Following recommendations from previous visual prompting studies of VLMs~\cite{redcircle,vipllava}, the boxes are red and have a line width equal to 1\% of the image size.

In separate-channel mask prompting, multiple deictic masks are represented in one channel by taking the pixelwise maximum regardless of the number of masks, $\mathbf M_t^{(m)}=\max_{\mathbf M\in\mathbb M_t^{(m)}}\mathbf M$.
The RGB observation remains unchanged, and $\mathbf M_t^{(m)}$ is supplied to the VLA as a mask prompt through an additional mask prompt fusion module.
The RGB image $\mathbf I$ and mask prompt $\mathbf M=\mathbf M_t^{(m)}$ are divided into $P\times P$ patches and independently projected into $D$ dimensions, as in a Vision Transformer~\cite{dosovitskiy2020vit}.
\begin{equation}
\begin{aligned}
\mathbf Z_I^{(0)}
&=
\operatorname{PatchEmbed}_I(\mathbf I)
+\mathbf E_{\mathrm{pos}},
\\
\mathbf Z_M
&=
\operatorname{PatchEmbed}_M(\mathbf M).
\end{aligned}
\end{equation}

\noindent The mask embeddings use no learnable bias, and therefore $\mathbf Z_M=\mathbf 0$ when $\mathbf M=\mathbf 0$. Following Alpha-CLIP~\cite{alphaclip}, we add $\mathbf Z_M$ elementwise to the image embeddings. We refer to the mask embeddings and this addition together as the mask prompt fusion module and design two variants with different fusion positions relative to the visual encoder.

In \textbf{MP-Early}, $\mathbf Z_M$ is added before the input to the visual encoder. In \textbf{MP-Late}, it is added after feature extraction by the $L$-layer visual encoder and before input to the LLM.
\begin{equation}
\begin{aligned}
\mathbf Z_{\mathrm{Early}}^{(0)}
&=
\mathbf Z_I^{(0)}+\mathbf Z_M,\\
\mathbf Z_{\mathrm{Late}}^{(L)}
&=
\mathcal V\!\left(\mathbf Z_I^{(0)}\right)
+\mathbf Z_M.
\end{aligned}
\end{equation}

\noindent Here, $\mathcal V$ denotes the visual encoder. For LI, $\mathbf M=\mathbf 0$ and therefore $\mathbf Z_M=\mathbf 0$, so neither MP-Early nor MP-Late changes the image embeddings. RGB visual prompting also renders no visual prompt for LI and uses the original RGB observation without modification. Thus, the zero mask in the canonical representation acts as a neutral input that leaves the visual input unchanged in either prompting family.

\subsection{Instruction-Mode Datasets and Two-Stage Training}

Let $\mathcal D_{\mathrm{LI}}$, $\mathcal D_{\mathrm{VLI}}$, and $\mathcal D_{\mathrm{VI}}$ denote the LI, VLI, and VI datasets, respectively. For each demonstration, we construct one variant for each of LI, VLI, and VI using the corresponding text prompt and deictic-mask input. We define the dataset containing all three modes as $\mathcal D_{\mathrm{all}}=\mathcal D_{\mathrm{LI}}\cup\mathcal D_{\mathrm{VLI}}\cup\mathcal D_{\mathrm{VI}}$ and the dataset containing only VLI and VI as $\mathcal D_{\mathrm{VLI+VI}}=\mathcal D_{\mathrm{VLI}}\cup\mathcal D_{\mathrm{VI}}$.

We denote full-parameter fine-tuning with initial parameters $\theta$ and dataset $\mathcal D$ by $\operatorname{FT}(\theta,\mathcal D)$. DeicticVLA uses two-stage training. We first obtain $\theta_{\mathrm{LI}}=\operatorname{FT}(\theta_{\mathrm{pre}},\mathcal D_{\mathrm{LI}})$ from the pretrained parameters $\theta_{\mathrm{pre}}$ and then train $\theta_{\mathrm{LI}\rightarrow\mathrm{all}}=\operatorname{FT}(\theta_{\mathrm{LI}},\mathcal D_{\mathrm{all}})$. To evaluate the contributions of the number of training stages and the inclusion of LI data in the second stage, we use single-stage training, $\theta_{\mathrm{pre}\rightarrow\mathrm{all}}=\operatorname{FT}(\theta_{\mathrm{pre}},\mathcal D_{\mathrm{all}})$, and training without LI data in the second stage, $\theta_{\mathrm{LI}\rightarrow\mathrm{VLI+VI}}=\operatorname{FT}(\theta_{\mathrm{LI}},\mathcal D_{\mathrm{VLI+VI}})$, as ablations. Section~\ref{sec:sim} details the composition of each dataset and the comparison conditions with the total number of training steps matched.


\section{Simulation Experiments}
\label{sec:sim}

Through simulation experiments on LIBERO~\cite{Liu2023LIBERO}, we compare RGB visual prompting and separate-channel mask prompting under two-stage training and ablate the number of training stages and the inclusion of LI data in the second stage. We address the following three questions:

\begin{enumerate}[label=Q\arabic*)]
  \item Under two-stage training, how do the four prompting methods differ in their in-distribution task success and their ability to use deictic masks in unseen layouts? 
  \item When the cumulative number of training steps is matched, is two-stage training more effective than single-stage training, and does its effect depend on the prompting method? 
  \item Does including LI data in the second stage mitigate forgetting of language instruction following while maintaining VLI and VI performance?
\end{enumerate}

\subsection{Experiment Setup}
\subsubsection{Tasks and Dataset Construction}

\begin{figure*}[t]
    \centering
    \includegraphics[width=.9\linewidth]{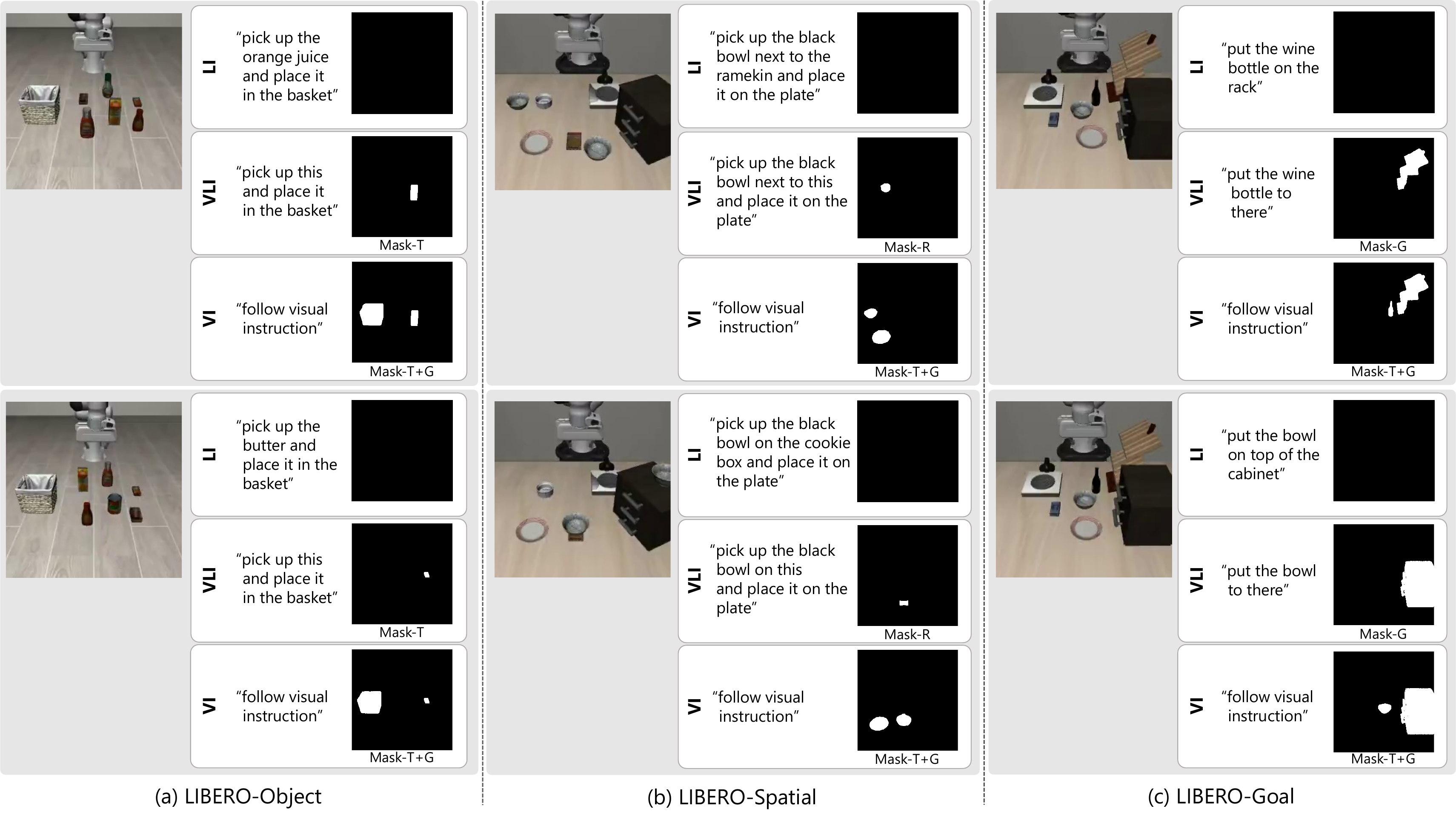}
    \caption{Example inputs for the three instruction modes in (a) LIBERO-Object, (b) LIBERO-Spatial, and (c) LIBERO-Goal. LI uses the original LIBERO text prompt. VLI uses Mask-T, Mask-R, or Mask-G depending on the suite. VI uses the fixed text prompt ``follow visual instruction'' with Mask-T+G, which consists of separate Mask-T and Mask-G deictic masks shown together in the same panel.}
    \label{fig:sim_tasks_overview}
\end{figure*}

Fig.~\ref{fig:sim_tasks_overview} shows the three LIBERO task suites and example inputs for each instruction mode. Each task contains 50 expert demonstrations. LI uses the original LIBERO text prompt. In VLI, Mask-T, Mask-R, and Mask-G specify the manipulated object, spatial reference object, and placement goal, respectively, depending on the suite. VI uses the fixed text prompt ``follow visual instruction'' and Mask-T+G, which indicates the manipulated object and placement goal. Mask-T+G consists of two deictic masks, Mask-T and Mask-G.

\textbf{LIBERO-Object} (Fig.~\ref{fig:sim_tasks_overview}(a)) consists of 10 tasks in which the robot grasps a target object from six visually distinct candidates and moves it to a basket. All tasks share the same six-object layout and differ only in the manipulated object.
\textbf{LIBERO-Spatial} (Fig.~\ref{fig:sim_tasks_overview}(b)) requires identifying one of two visually identical black bowls based on its spatial relation to a reference object and moving it to a plate. We use 9 of the 10 tasks, excluding the task in which the target bowl is at the center of the table and no single reference object can be defined.
\textbf{LIBERO-Goal} (Fig.~\ref{fig:sim_tasks_overview}(c)) requires moving a specified object to a specified goal location under a fixed object set and layout. We use only the six ``put A B'' tasks that include both a manipulated object and a placement goal.

To eliminate the effect of segmentation accuracy when comparing prompting methods and training strategies, we generate deictic masks from the simulator's ground-truth instance segmentation rather than from SAM~2.
\begin{table}[t]
    \caption{Training Hyperparameters\label{tab:training_hyperparameters}}
    \centering
    \begin{tabular}{ll}
        \toprule
        Optimizer & AdamW ($\beta_1=0.9$, $\beta_2=0.95$, $\epsilon=10^{-8}$) \\
        Weight decay & $10^{-10}$ \\
        Gradient clipping & $\|\mathbf{g}\|_2 \leq 1.0$ \\
        LR schedule & Cosine decay (warmup: 1k steps) \\
        Learning rate & $2.5 \times 10^{-5} \to 2.5 \times 10^{-6}$ \\
        Total training steps & 60{,}000 \\
        Batch size & 32 \\
        EMA decay rate & 0.99 \\
        \bottomrule
    \end{tabular}
\end{table}

\begin{table*}[t]
  \centering
  \caption{Simulation results. Mean SR is averaged over LI, VLI, and VI. Higher $\Delta$SR indicates greater contribution from deictic masks.}
  \label{tab:combined_sim_results}
  \footnotesize
  \renewcommand{\arraystretch}{0.96}
  \setlength{\tabcolsep}{7.0pt}
  \begin{tabular}{@{}ll rrrr rr rr rr@{}}
    \toprule
    & & \multicolumn{4}{c}{In-distribution SR}
    & \multicolumn{2}{c}{\makebox[0pt][c]{In-distribution $\Delta\mathrm{SR}$}}
    & \multicolumn{2}{c}{\makebox[0pt][c]{Object-ZS $\Delta\mathrm{SR}$}}
    & \multicolumn{2}{c@{}}{\makebox[0pt][c]{Spatial-ZS $\Delta\mathrm{SR}$}} \\
    \cmidrule(lr){3-6} \cmidrule(lr){7-8} \cmidrule(lr){9-10} \cmidrule(lr){11-12}
    Training & Prompting & LI & VLI & VI & Mean
    & \makebox[2.8em][c]{VLI} & \makebox[2.8em][c]{VI}
    & \makebox[2.8em][c]{VLI} & \makebox[2.8em][c]{VI}
    & \makebox[2.8em][c]{VLI} & \makebox[2.8em][c]{VI} \\
    \midrule
    \multirow{4}{*}{2S}
      & MP-Early & 93.0 & 96.3 & 96.9 & 95.4 & +23.1 & +31.7 & +9.1 & +10.0 & +3.1 & +4.4 \\
      & MP-Late  & 91.5 & 95.5 & 95.3 & 94.1 & +22.3 & +33.4 & \textbf{+29.2} & +25.2 & +0.9 & +1.7 \\
      & VP-Fade  & 93.1 & 96.2 & 95.6 & 95.0 & \textbf{+25.6} & \textbf{+35.2} & +26.2 & +24.8 & +1.3 & +5.8 \\
      & VP-BBox  & 94.0 & 96.8 & 96.1 & \underline{\textbf{95.6}} & \textbf{+25.6} & +34.0 & \textbf{+29.2} & \textbf{+27.8} & \textbf{+16.0} & \textbf{+11.1} \\
    \midrule
    \multirow{4}{*}{1S}
      & MP-Early & 90.5 & 92.3 & 91.4 & 91.4 & +19.1 & +24.1 & +3.5 & +3.4 & +0.2 & +0.9 \\
      & MP-Late  & 91.8 & 95.2 & 94.1 & 93.7 & +21.5 & +31.2 & \textbf{+7.2} & \textbf{+8.0} & \textbf{+1.3} & $-$0.5 \\
      & VP-Fade  & 93.6 & 96.9 & 96.0 & \textbf{95.5} & \textbf{+24.6} & \textbf{+33.2} & +0.7 & +0.2 & +0.4 & +0.0 \\
      & VP-BBox  & 92.4 & 91.9 & 94.0 & 92.8 & +22.1 & +28.0 & +0.9 & +1.0 & +1.2 & \textbf{+2.2} \\
    \midrule
    \multirow{4}{*}{2S-NoLI}
      & MP-Early & 64.6 & 96.9 & 96.2 & 85.9 & +21.5 & +27.8 & +16.5 & +17.9 & +0.0 & $-$0.1 \\
      & MP-Late  & 63.6 & 96.4 & 95.7 & 85.2 & +20.9 & +30.6 & +9.3 & +10.3 & +0.4 & +0.5 \\
      & VP-Fade  & 69.2 & 97.0 & 97.0 & 87.7 & \textbf{+25.4} & \textbf{+32.9} & \textbf{+26.5} & \textbf{+25.7} & $-$0.5 & \textbf{+4.0} \\
      & VP-BBox  & 72.1 & 96.0 & 95.4 & \textbf{87.8} & +22.5 & +29.7 & +18.8 & +19.6 & \textbf{+3.5} & +3.6 \\
    \bottomrule
  \end{tabular}
\end{table*}

\subsubsection{Zero-Shot Evaluation Suites}
To evaluate generalization to unseen layouts, we construct two evaluation-only suites from LIBERO-Object and LIBERO-Spatial. \textbf{LIBERO-Object-ZS} consists of four sub-suites that progressively perturb the initial object arrangement of LIBERO-Object so that the task cannot be identified from the initial observation alone. Object-dist-shuffle randomly permutes the positions of the five distractors while keeping the target fixed. Object-target-swap exchanges the target position with that of one other object. Object-swap-shuffle combines these two perturbations. Object-full-shuffle randomly permutes all six objects without distinguishing the target from the distractors. In \textbf{LIBERO-Spatial-ZS}, all eight black-bowl positions used across the original LIBERO-Spatial suite are populated simultaneously, whereas each original task contains only the target bowl and one other bowl. This construction makes the bowl arrangement identical across tasks and prevents task identification from the initial observation alone.

\subsubsection{Compared Configurations and Evaluation Metrics}
We compare the four methods defined in Section~\ref{subsec:vla_and_robot_component}, namely VP-Fade, VP-BBox, MP-Early, and MP-Late, using two-stage training as the main condition. For each method, we further ablate the number of training stages and the inclusion of LI data in the second stage, yielding 12 conditions in total. We denote two-stage training and the two ablations by the following abbreviations.
\begin{equation}
\mathrm{2S}\equiv\theta_{\mathrm{LI}\rightarrow\mathrm{all}}, \ 
\mathrm{2S\text{-}NoLI}\equiv\theta_{\mathrm{LI}\rightarrow\mathrm{VLI+VI}}, \ 
\mathrm{1S}\equiv\theta_{\mathrm{pre}\rightarrow\mathrm{all}}.
\end{equation}

\noindent Here, NoLI in 2S-NoLI indicates that the second-stage training excludes LI data. The first stage uses $\mathcal D_{\mathrm{LI}}$, as in 2S.

We evaluate each task over 25 trials and use the task success rate, denoted by SR. For VLI and VI, we define the contribution of deictic masks to task success as
\begin{equation}
\Delta\mathrm{SR}
=
\mathrm{SR}_{\mathrm{deictic}}
-
\mathrm{SR}_{\mathrm{no\text{-}deictic}}.
\end{equation}

\noindent $\mathrm{SR}_{\mathrm{deictic}}$ is measured using inputs containing valid deictic masks. $\mathrm{SR}_{\mathrm{no\text{-}deictic}}$ is measured while retaining the text prompt and disabling only the input derived from the deictic masks. For MP, the latter uses a zero mask, whereas for VP it uses the unmodified RGB observation without a rendered visual prompt.

\subsubsection{Training Details}
We use $\pi_0$~\cite{pi0} as the base pretrained VLA and fine-tune all parameters. We train one model for each configuration. For 2S and 2S-NoLI, the first stage on $\mathcal D_{\mathrm{LI}}$ consists of 30k steps, followed by 30k steps in the second stage. For 1S, the model is trained on $\mathcal D_{\mathrm{all}}$ for 60k steps. Thus, the cumulative number of training steps from $\theta_{\mathrm{pre}}$ is fixed at 60k for all conditions. Table~\ref{tab:training_hyperparameters} lists the shared hyperparameters, including the optimizer, learning rate, and batch size. Training uses 2$\times$A100 GPUs.

\subsection{Results}
Table~\ref{tab:combined_sim_results} summarizes SR and $\Delta\mathrm{SR}$ across the in-distribution and zero-shot evaluations. We answer the three questions based on these results.

\subsubsection{Prompting Method Comparison under Two-Stage Training (Q1)}

All prompting methods achieve mean SRs of 94.1--95.6\% in the in-distribution evaluation. Although VP-BBox attains the highest value of 95.6\%, the maximum difference among the methods is only 1.5 points, indicating that task success is largely saturated. For $\Delta\mathrm{SR}$, VP-Fade reaches $+25.6$ points for VLI and $+35.2$ points for VI, while VP-BBox reaches $+25.6$ and $+34.0$ points, respectively. These two RGB visual prompting methods attain the highest values. MP-Late reaches $+22.3$ and $+33.4$ points, and MP-Early reaches $+23.1$ and $+31.7$ points, showing contributions from deictic masks close to those of visual prompting.

The differences among methods are clearer in the zero-shot evaluation. On Object-ZS, VP-BBox reaches $+29.2$ points for VLI and $+27.8$ points for VI. MP-Late also reaches $+29.2$ and $+25.2$ points, respectively, performing comparably to VP-BBox. VP-Fade scores slightly below these two methods, whereas MP-Early performs substantially worse than the other three. On Spatial-ZS, VP-BBox reaches $+16.0$ and $+11.1$ points, while all other methods remain in the single digits. The other methods have lower $\mathrm{SR}_{\mathrm{deictic}}$ because they more often select a visually identical distractor bowl despite valid masks, whereas VP-BBox is less affected by this failure mode.

Overall, VP-BBox achieves the highest performance under 2S and shows a clear advantage particularly on Spatial-ZS. MP-Late achieves the highest zero-shot performance among the separate-channel mask prompting methods and performs comparably to VP-BBox on Object-ZS.

\subsubsection{Training-Stage Ablation (Q2)}

The difference between 2S and 1S in the in-distribution evaluation varies by prompting method. The mean SRs of MP-Late and VP-Fade remain nearly unchanged, whereas those of MP-Early and VP-BBox decrease by 4.0 and 2.8 points, respectively, under 1S. All prompting methods retain relatively high in-distribution SR and $\Delta\mathrm{SR}$ under 1S, making the effect of the number of training stages unclear from the in-distribution evaluation alone.

On Object-ZS, however, 1S reduces $\Delta\mathrm{SR}$ for all four evaluated methods. In particular, the values for VP-Fade and VP-BBox fall to nearly zero. Although MP-Late also performs worse than under 2S, it retains $+7.2$ points for VLI and $+8.0$ points for VI, the highest performance under 1S. On Spatial-ZS, the improvement achieved by VP-BBox under 2S also nearly disappears under 1S.

Thus, when the cumulative number of training steps is matched, 2S is effective for acquiring the ability to use deictic masks in unseen layouts rather than for improving in-distribution task success. Its effect depends on the prompting method. RGB visual prompting depends strongly on two-stage training, whereas MP-Late retains some zero-shot performance under 1S.

\begin{figure*}[t]
    \centering
    \includegraphics[width=\linewidth]{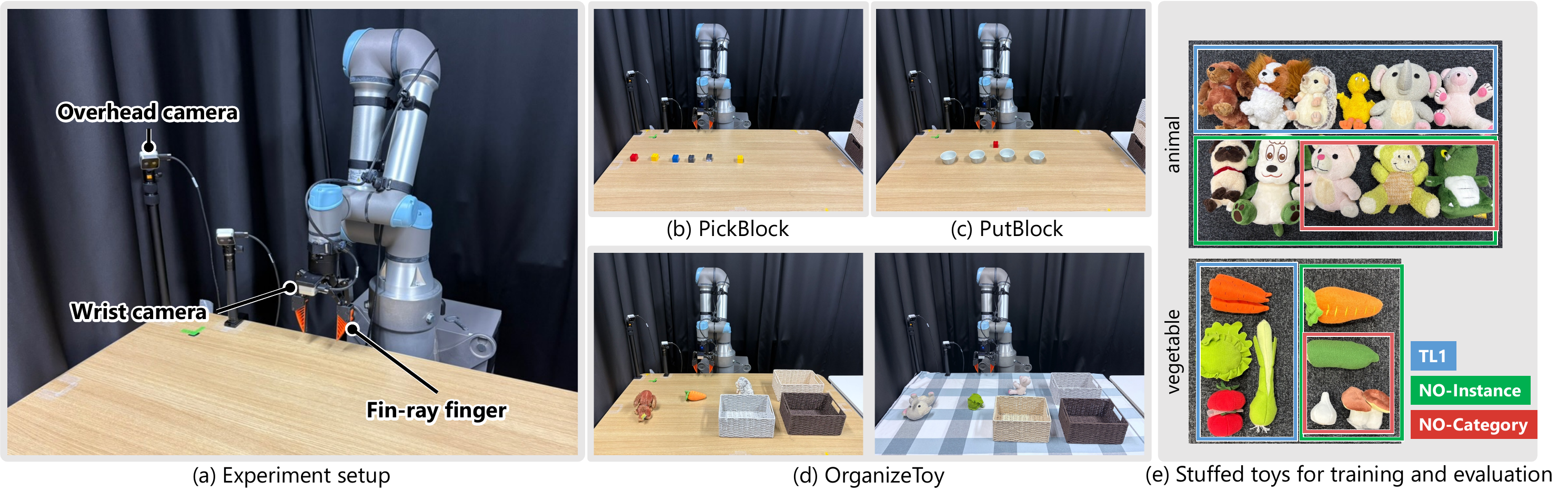}
    \caption{Real-world experiment setup and task environments. (a) Hardware setup with a UR5e, Robotiq 2F-85 with Fin-ray fingers, overhead camera, and wrist camera. (b) \textbf{PickBlock}: pick up a specified block from six colored blocks on the table. (c) \textbf{PutBlock}: put a block into a specified bowl among visually identical white bowls arranged in a row. (d) \textbf{OrganizeToy}: pick up a specified stuffed toy and place it into a specified basket; left shows the standard setting, right shows the VC-Surface condition with a tablecloth not seen during training. (e) Stuffed toys used in OrganizeToy across two super-categories: animal and vegetable. TL1 denotes seen objects used for both training and evaluation, while NO-Instance and NO-Category denote novel objects used only for evaluation.}
    \label{fig:real_exp_setup}
\end{figure*}

\subsubsection{Second-Stage LI-Data Ablation (Q3)}

Under 2S-NoLI, VLI and VI maintain high SRs of 95.4--97.0\%, whereas LI SRs decrease to 63.6--72.1\%. Relative to 2S, LI decreases by 21.9--23.9 points for VP-Fade and VP-BBox and by 27.9--28.4 points for MP-Early and MP-Late, demonstrating forgetting of language instruction following for every method. Meanwhile, $\Delta\mathrm{SR}$ under 2S-NoLI remains at $+20.9$--$+25.4$ points for VLI and $+27.8$--$+32.9$ points for VI. Thus, excluding LI data from the second stage allows the model to acquire the ability to use deictic masks but selectively degrades its ability to handle LI.

The effect on zero-shot evaluation varies by prompting method. Object-ZS performance remains nearly unchanged for VP-Fade and improves for MP-Early, whereas it decreases for MP-Late and VP-BBox. In particular, both MP-Late and VP-BBox, which perform well on Object-ZS under 2S, deteriorate, and the improvement of VP-BBox on Spatial-ZS is also substantially reduced under 2S-NoLI. Therefore, excluding LI data from the second stage provides no consistent improvement in zero-shot performance and instead degrades the generalization of methods that are promising under 2S.

Overall, jointly training LI, VLI, and VI in the second stage preserves language instruction following without impairing the ability to use deictic masks in VLI and VI. It is also effective in maintaining the zero-shot performance of the promising prompting methods, VP-BBox and MP-Late.


\section{Real-World Experiments}
\label{sec:real_experiments}

Using three real-world manipulation tasks, we investigate whether DeicticVLA can support LI, VLI, and VI within a single policy learned from small-scale real-world data. We further evaluate the effectiveness of VLI and VI for instruction expressions unseen during training, visual perturbations, and novel objects by addressing the following four questions:

\begin{enumerate}[label=Q\arabic*)]
  \item Can a single policy supporting LI, VLI, and VI be learned from small-scale real-world data, and what task success does each instruction mode achieve compared with a policy trained only with language instructions?
  \item Do VLI and VI achieve higher task success than LI for instructions containing combinations or expressions unseen during training?
  \item Do VLI and VI maintain their advantage over LI when the appearance of the workspace differs from that during training?
  \item Do VLI and VI achieve higher task success than LI for object instances and categories unseen during training?
\end{enumerate}

\subsection{Experiment Setup}
\label{subsec:real_setup}

\begin{table*}[t]
  \caption{Task templates for each task and condition.}
  \label{tab:instruction_patterns}
  \centering
  \begin{tabular}{cl@{\hskip 1.5em}l@{\hskip 1.5em}l}
    \toprule
    TL  & PickBlock & PutBlock & OrganizeToy \\
    \midrule
    \midrule
    TL1 & \textit{pick up \ldots}
        & \textit{put the block into \ldots}
        & \textit{put the \ldots}  \\
        & \quad the \{yellow, black, blue, red\} block,
        & \quad the \{1st, 2nd\} bowl from the right ($N{=}2\text{--}5$),
        & \quad \{animal, vegetable\} into the \\
        & \quad the \{1st, 2nd, 3rd, 4th\} block from the left,
        & \quad the 3rd bowl from the right ($N{=}3\text{--}5$).
        & \qquad \{front, back left, back right\} basket, \\
        & \quad the \{yellow, blue\} block on the right,
        &
        & \quad \textless{}category\textgreater{} into the \\
        & \quad the \{black, red\} block on the left.
        &
        & \qquad \{front, back left, back right\} basket. \\
    \midrule
    TL2 & \textit{pick up \ldots}
        & \textit{put the block into \ldots}
        & \\
        & \quad the \{yellow, blue\} block on the left,
        & \quad the \{1st, 2nd\} bowl from the left ($N{=}2\text{--}5$),
        & \\
        & \quad the \{red, black\} block on the right,
        & \quad the 3rd bowl from the left ($N{=}3\text{--}5$).
        & \\
        & \quad the \{1st, 2nd, 3rd, 4th\} block from the right.
        &
        & \\
    \midrule
    TL3 & \textit{pick up \ldots}
        & \textit{put the block into \ldots}
        & \\
        & \quad the \{green, orange\} block,
        & \quad the 4th bowl from the \{left, right\} ($N{=}4{,}5$),
        & \\
        & \quad the \{leftmost, rightmost\} block,
        & \quad the 5th bowl from the \{left, right\} ($N{=}5$),
        & \\
        & \quad the \{5th, 6th\} block from the \{left, right\}.
        & \quad the \{leftmost, rightmost\} bowl ($N{=}2\text{--}5$),
        & \\
        &
        & \quad the center bowl ($N{=}3{,}5$).
        & \\
    \bottomrule
  \end{tabular}
\end{table*}

\begin{table}[t]
  \centering
  \caption{Stuffed toys used in OrganizeToy. For entries containing parentheses, the toy name appears first, followed by the category label used in LI in parentheses.}
  \label{tab:toy_objects}
  \begin{tabular}{lll}
    \toprule
    Condition & Animal & Vegetable \\
    \midrule
    \midrule
    TL1
      & dachshund (dog),
      & carrot-bunch (carrot), \\
      & chihuahua (dog),
      & lettuce, green onion, \\
      & hedgehog, bird, elephant, pig.
      & tomato. \\
    \midrule
    NO-Instance
      & bulldog (dog), wanwan (dog),
      & carrot-single (carrot), \\
      & pink bear (animal),
      & cucumber (vegetable), \\
      & monkey (animal),
      & mushroom (vegetable), \\
      & crocodile (animal).
      & onion (vegetable). \\
    \midrule
    NO-Category
      & pink bear (bear), monkey,
      & cucumber, mushroom, \\
      & crocodile, mouse.
      & onion. \\
    \bottomrule
  \end{tabular}
\end{table}

\subsubsection{Hardware Setup}
Fig.~\ref{fig:real_exp_setup}(a) shows the real-world experiment setup. A Robotiq 2F-85 gripper equipped with Fin-ray fingers fabricated based on UMI~\cite{umi} is mounted on a UR5e. RGB observations are obtained from an overhead camera that captures the entire table and a wrist camera fixed to the gripper. The user clicks the target on a tablet displaying the overhead camera image, and SAM~2~\cite{sam2} generates a deictic mask. After generation, the tracking function of SAM~2 updates the mask so that it follows the target object during manipulation. When both Mask-T and Mask-G are used, the user generates them by clicking the corresponding objects sequentially.

\subsubsection{Tasks}
We designed the three tasks shown in Fig.~\ref{fig:real_exp_setup}(b)--(d), namely PickBlock, PutBlock, and OrganizeToy. PickBlock and PutBlock evaluate generalization to unseen instruction expressions, whereas OrganizeToy evaluates generalization to changes in workspace appearance and novel objects. Table~\ref{tab:instruction_patterns} lists the task templates for each task and condition.

\textbf{PickBlock} requires the robot to grasp one specified block from six colored blocks placed on the table. Because blocks of the same color are present, LI specifies the target by its color, its left-right relation, or its ordinal position counted from the left or right. VLI uses ``pick up this block'' with Mask-T, whereas VI uses Mask-T.

\textbf{PutBlock} requires the robot to place a single block into a specified goal selected from two to five visually identical white bowls arranged in a row. LI specifies the bowl using an ordinal position counted from the left or right, or a spatial expression such as leftmost, center, or rightmost. VLI uses ``put the block into this bowl'' with Mask-G, whereas VI uses Mask-T+G.

\textbf{OrganizeToy} requires the robot to grasp a specified stuffed toy and place it into a specified basket. LI specifies the task using a super-category such as animal or vegetable, or the category label of each object and the basket position. The objects and LI category labels used in OrganizeToy are listed in Table~\ref{tab:toy_objects}. VLI uses ``put this into there'' with Mask-T+G, whereas VI uses Mask-T+G.

\begin{table*}[t]
  \centering
  \caption{Success rates in real-world experiments. Values outside and inside parentheses denote success rates (\%) and numbers of successful trials, respectively; $n$ denotes the total number of trials in each column.}
  \label{tab:real_results}
  \begin{tabular}{l rrr rrr rrrr}
    \toprule
    & \multicolumn{3}{c}{PickBlock}
    & \multicolumn{3}{c}{PutBlock}
    & \multicolumn{4}{c}{OrganizeToy} \\
    \cmidrule(lr){2-4} \cmidrule(lr){5-7} \cmidrule(lr){8-11}
    \shortstack[l]{Instruction\\mode}
    & TL1            & TL2            & TL3
    & TL1            & TL2            & TL3
    & TL1            & VC-Surface     & NO-Instance    & NO-Category \\
    & ($n{=}24$)     & ($n{=}16$)     & ($n{=}16$)
    & ($n{=}15$)     & ($n{=}15$)     & ($n{=}20$)
    & ($n{=}40$)     & ($n{=}40$)     & ($n{=}24$)     & ($n{=}12$) \\
    \midrule\midrule
    LI (baseline)
    & 70.8 (17)            & 12.5 (\phantom{0}2)  & 18.8 (\phantom{0}3)
    & 80.0 (12)            & \phantom{0}6.7 (\phantom{0}1) & 30.0 (\phantom{0}6)
    & 67.5 (27)            & 65.0 (26)            & 66.7 (16)            & 33.3 (\phantom{0}4) \\
    LI
    & 29.2 (\phantom{0}7)  & 12.5 (\phantom{0}2)  & 18.8 (\phantom{0}3)
    & 40.0 (\phantom{0}6)  & 26.7 (\phantom{0}4)  & 20.0 (\phantom{0}4)
    & 62.5 (25)            & 60.0 (24)            & 62.5 (15)            & 16.7 (\phantom{0}2) \\
    VLI
    & 50.0 (12)            & 62.5 (10)            & 68.8 (11)
    & 73.3 (11)            & 86.7 (13)            & 80.0 (16)
    & 77.5 (31)            & 87.5 (35)            & 95.8 (23)            & \phantom{0.}100 (12) \\
    VI
    & 45.8 (11)            & 62.5 (10)            & 93.8 (15)
    & 53.3 (\phantom{0}8)  & 86.7 (13)            & 75.0 (15)
    & 75.0 (30)            & 90.0 (36)            & 95.8 (23)            & \phantom{0.}100 (12) \\
    \bottomrule
  \end{tabular}
\end{table*}

\subsubsection{Experiment Conditions}
We define the following experiment conditions, which are applied across the tasks described above.

\textbf{Task Levels.} 
Following IA-VLA~\cite{Hannus2025IAVLA}, we divide the PickBlock and PutBlock instructions into TL1--TL3 according to their relationship to the training distribution. TL1 is an in-distribution condition consisting of target specifications used during training. TL2 is an out-of-distribution (OoD) condition containing unseen combinations of learned concepts, such as reversing the direction in which ordinal positions are counted. TL3 is an OoD condition requiring extrapolation to unseen concepts, including colors, spatial expressions, and ordinal positions not observed during training.

For PickBlock, TL1 contains color and position expressions used during training, TL2 contains combinations of known colors and unseen position expressions, and TL3 contains unseen colors, spatial expressions, and ordinal positions. For PutBlock, TL1 contains ordinal positions counted from the right as used during training, TL2 contains ordinal positions counted from the left, which is an unseen counting direction, and TL3 contains unseen ordinal positions and spatial expressions.

\textbf{Visual Change.} 
To evaluate robustness to visual perturbations, we define the VC-Surface condition for OrganizeToy. In this condition, a tablecloth not used during training is placed on the table to change the appearance of its surface.

\textbf{Novel Objects.} 
To evaluate generalization to objects unseen during training, we define the NO-Instance and NO-Category conditions for OrganizeToy. NO-Instance uses novel instances belonging to categories used during training. NO-Category uses objects from categories not included in the training data. The object assignments for TL1, NO-Instance, and NO-Category are summarized in Table~\ref{tab:toy_objects}.

\subsubsection{Data Collection and Training}
Demonstration data were collected at 30~Hz using GELLO~\cite{gello}, a joint-angle-based teleoperation device. Each episode records overhead and wrist camera images, deictic masks, proprioception, gripper state, and actions. For PickBlock, we collected 100 episodes using the TL1 instruction templates. For PutBlock, we collected 110 episodes using the TL1 instruction templates while varying the number and arrangement of bowls. For OrganizeToy, we collected 10 episodes using a category name and 5 episodes using a super-category name for each training object, for a total of 150 episodes. In total, the collected demonstration data comprise 360 episodes and approximately 97k timesteps.

We use $\pi_0$~\cite{pi0} as the base VLA. Although VP-BBox performed best under two-stage training, we selected MP-Late for the real-world evaluation because it retained zero-shot instruction following relatively well across training strategies and leaves the RGB observation unchanged. This avoids potential visual interference from the red bounding box used by VP-BBox and potential loss of contextual information caused by the background attenuation used by VP-Fade. We combine the data from the three tasks and train on $\mathcal D_{\mathrm{LI}}$ for 15k steps in the first stage and on $\mathcal D_{\mathrm{all}}$ for 15k steps in the second stage. For comparison, we use a language-instruction-only policy trained from $\pi_0$ on $\mathcal D_{\mathrm{LI}}$ for 30k steps as LI~(baseline). All other training conditions are identical to those in Table~\ref{tab:training_hyperparameters}.

\begin{figure}[t]
  \centering
  \includegraphics[width=\linewidth]{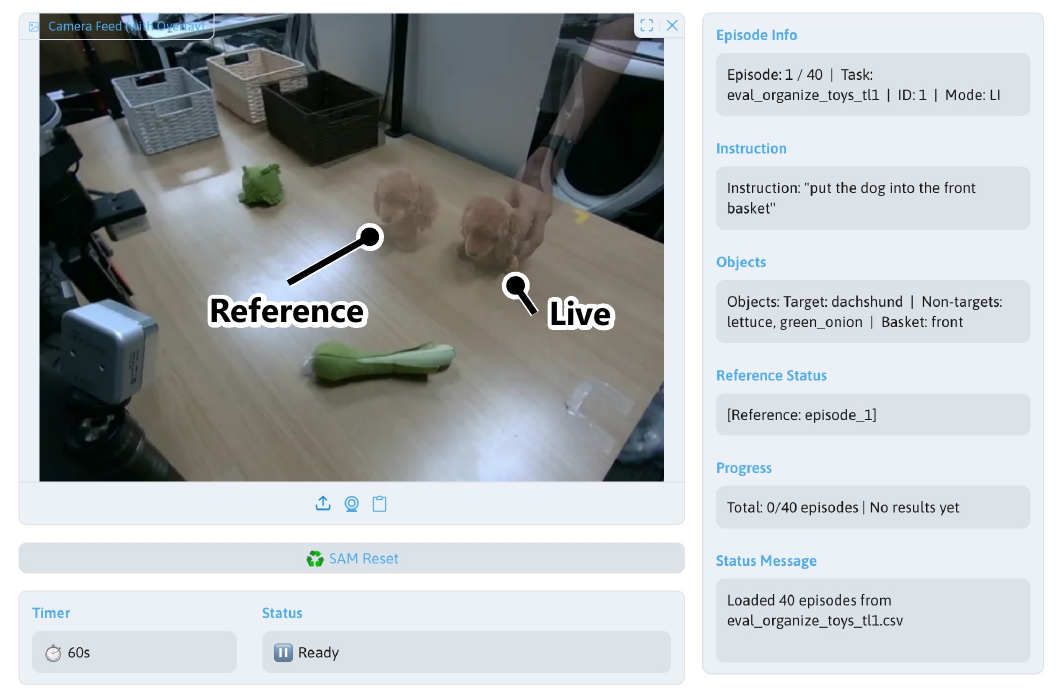}
  \caption{Evaluation GUI for aligning the initial scene configuration. Before each trial, the evaluator adjusts the objects so that their positions in the live overhead-camera image (Live) match those in the overlaid reference image (Reference).}
  \label{fig:evaluation_gui}
\end{figure}

\subsubsection{Evaluation Protocol}
To support the real-world experiments, we developed the evaluation GUI shown in Fig.~\ref{fig:evaluation_gui}. Variations in the initial scene configuration can confound real-world policy evaluation because object poses cannot be reset as exactly as in simulation. To mitigate this issue, inspired by the initial-condition overlay procedure of the TRI LBM Team~\cite{tri_lbm}, the GUI overlays a captured reference image on the live overhead-camera image. Before each evaluation trial, the evaluator manually adjusted the objects until their positions and orientations in the live image matched those in the reference overlay, as illustrated in Fig.~\ref{fig:evaluation_gui}. This procedure reduces operator-dependent variation in the initial conditions and improves consistency across trials.

\FloatBarrier

\subsection{Results}
\label{subsec:real_results}

Table~\ref{tab:real_results} shows the success rate for each task, instruction mode, and experimental condition. The number of evaluation trials is shown as $n$ for each column.

\subsubsection{Effectiveness at Training-Distribution Tasks (Q1)}

VLI and VI outperformed LI in all three tasks. VLI achieved 50.0--77.5\% and VI achieved 45.8--75.0\%, whereas LI achieved 29.2--62.5\%. Thus, a single policy supporting all three instruction modes was learned from 100--150 episodes per task.

In comparison with LI~(baseline), however, LI performance decreased after joint training. The success rate decreased from 70.8\% to 29.2\% in PickBlock and from 80.0\% to 40.0\% in PutBlock. The difference was smaller in OrganizeToy, where the corresponding success rates were 67.5\% and 62.5\%. This difference may be related to the fact that PickBlock and PutBlock require the interpretation of detailed ordinal and spatial expressions, whereas OrganizeToy can be solved using category-level specifications. A lower sampling ratio for LI data in the second stage may also have contributed to the decrease. These results show that MP-Late with two-stage training realizes LI, VLI, and VI within a single policy, while leaving room to improve the retention of language instruction following.

PickBlock had lower success rates than the other tasks across all instruction modes. In failure cases, the policy tended to grasp a block that appeared large in the wrist camera rather than the instructed target. Because the blocks and deictic masks appear small in the overhead camera, reliance on the wrist camera may have interfered with target specification. However, preliminary overhead-only trials on both block tasks suffered from inaccurate reaching, motivating improved multi-view fusion rather than wrist-camera removal.

\subsubsection{Generalization to Unseen Instructions (Q2)}

At TL2 and TL3 in PickBlock and PutBlock, LI and LI~(baseline) decreased substantially from TL1, whereas VLI and VI maintained high success rates. In PickBlock, LI and LI~(baseline) both achieved 12.5\% at TL2 and 18.8\% at TL3. In contrast, VLI achieved 62.5\% and 68.8\%, and VI achieved 62.5\% and 93.8\%, respectively. In PutBlock, LI~(baseline) achieved 6.7\% at TL2 and 30.0\% at TL3, while LI achieved 26.7\% and 20.0\%. In contrast, VLI achieved 86.7\% and 80.0\%, and VI achieved 86.7\% and 75.0\%, respectively.

These results show that generalization to unseen counting directions, ordinal positions, and spatial expressions is difficult with language alone. By directly specifying the referent using deictic masks, VLI and VI reduce their dependence on instance-level grounding through language. Notably, VLI and VI tended to achieve higher success rates at TL2 and TL3 than at TL1. One possible explanation is that, in TL1, which follows the training distribution, the model overfits to correlations among specific objects, scene configurations, and actions in the training data, and this bias interferes with target specification using mask prompts. In TL2 and TL3, which contain unseen combinations or concepts, these learned correlations are less readily available. The model may therefore rely more strongly on mask prompts, resulting in higher success rates.

\subsubsection{Robustness to Visual Perturbations (Q3)}

Under VC-Surface in OrganizeToy, LI~(baseline) and LI achieved 65.0\% and 60.0\%, respectively, which did not differ substantially from their TL1 success rates of 67.5\% and 62.5\%. Thus, changing only the table surface did not substantially impair language instruction following. In contrast, VLI and VI achieved 87.5\% and 90.0\%, respectively, and substantially outperformed LI as they did at TL1. This result shows that the advantage of VLI and VI was maintained under at least the appearance change used in this experiment.

\subsubsection{Generalization to Novel Objects (Q4)}

Under NO-Instance in OrganizeToy, LI~(baseline) and LI achieved 66.7\% and 62.5\%, respectively, whereas VLI and VI both achieved 95.8\%. Thus, VLI and VI with MP-Late maintained high task success for novel instances belonging to known categories.

Under NO-Category, LI~(baseline) achieved 33.3\% and LI achieved only 16.7\%, whereas both VLI and VI achieved 100\%. This result means that the task could be completed by specifying the manipulation target and placement goal with deictic masks without using category names in language. It therefore suggests the usefulness of VLI and VI in real environments where novel objects appear frequently.


\section{Discussion}

\subsection{Benefits and Challenges of Supporting Three Instruction Modes within a Single VLA}

\subsubsection{Complementary Roles of LI, VLI, and VI}

The significance of supporting LI, VLI, and VI lies not in using any one instruction mode at all times, but in making complementary means of target specification available according to the task and environment.

LI is useful in two respects. First, it can enable hands-free operation when language is provided through speech. Second, abstract or quantified expressions such as ``all the red blocks'' and ``any vegetable'' cannot be expressed naturally using deictic masks alone.

In contrast, VLI and VI are useful when target identification through language is unreliable. In the real-world experiments, VLI and VI compensated for failures of LI under conditions involving unseen ordinal and spatial expressions as well as novel objects. The three modes are therefore not substitutes for one another. Their value lies in selecting among them according to the nature of the task, the context of use, and the environment.

\subsubsection{Flexibility and Scalability of a Single Policy}

Supporting the three instruction modes within a single policy allows the user to make this selection without switching the underlying policy. For example, the user can rely on LI in ordinary use and fall back to VLI or VI within the same task when the target proves difficult to identify through language, or choose a mode simply according to preference.

Similar flexibility could be provided by preparing a separate policy for each instruction mode and selecting among them. However, each mode-specific policy would then require its own fine-tuning, evaluation, and deployment; this burden is not negligible when each policy is obtained by full-parameter fine-tuning of a large pretrained VLA, as in this study, and it grows with the number of instruction modes and target tasks. In contrast, DeicticVLA absorbs the differences among the modes through instruction canonicalization, so only a single policy needs to be fine-tuned and maintained, providing a more scalable way to support multiple instruction modes.

\subsubsection{Balancing Instruction Modes during Joint Training}

The simulation experiments showed that including LI data during the second stage of joint training is important for preserving language instruction following while acquiring the ability to follow VLI and VI. We also applied this training recipe to the real-world experiments, but LI after joint training underperformed the LI-only baseline on PickBlock and PutBlock. Thus, although including LI data in joint training is important, the sampling ratio for LI data requires further adjustment when using small-scale real-world datasets.

\subsection{Modularity and Extensibility of DeicticVLA}

\subsubsection{Prompting Method as an Implementation Choice}

DeicticVLA separates the instruction mode selected by the user from the prompting method used to provide deictic masks to the VLA. LI, VLI, and VI are converted into a common representation comprising a text prompt and deictic masks, after which either RGB visual prompting or separate-channel mask prompting is applied. The unified three-mode interface of DeicticVLA is therefore not limited to a specific prompting method or base VLA.

The simulation experiments showed no consistent superiority of RGB visual prompting over separate-channel mask prompting or vice versa. The four methods achieved similar performance within the training distribution. VP-BBox achieved the highest performance on Spatial-ZS, whereas MP-Late retained some ability to follow Object-ZS instructions even under single-stage training. These results indicate that the prompting method should be regarded not as a fixed component of DeicticVLA, but as an implementation choice that can be selected according to the target VLA, the required generalization conditions, and the operating environment.

We adopted MP-Late for the real-world system because it retained zero-shot instruction following relatively well across changes in the training strategy, left the RGB observation unmodified, and used the masks generated and tracked by SAM~2 without additional image-rendering operations. The real-world experiments demonstrated the applicability of DeicticVLA with MP-Late to a physical robot. However, because the other prompting methods were not evaluated on the physical robot, these experiments do not demonstrate the superiority of mask prompting in real-world settings.

\subsubsection{Applicability to Other Base VLAs}

Because DeicticVLA does not change the action space or training objective of the base VLA, it can be applied to pretrained VLAs other than $\pi_0$. In particular, RGB visual prompting can be introduced without modifying the architecture of the base VLA, whereas separate-channel mask prompting can be introduced by adding a mask input pathway. DeicticVLA also uses the visual recognition, language understanding, and action generation capabilities of the base VLA. Performance across the three instruction modes is therefore expected to improve as more capable base VLAs become available.

\subsection{Limitations and Future Directions for Human--Robot Interaction}

\subsubsection{Mask Acquisition and Real-World Evaluation}

The simulation experiments used ground-truth segmentation masks to isolate the effect of the prompting method itself. In contrast, the real-world experiments generated and tracked masks from user clicks using SAM~2. This study therefore demonstrated the feasibility of a real-world pipeline that includes SAM~2, but did not separately evaluate the effects of segmentation and tracking errors on policy performance.

Furthermore, only MP-Late was evaluated on the physical robot, and RGB visual prompting and separate-channel mask prompting were not compared in the real world. Separate-channel mask prompting has the structural properties of preserving the RGB observation and avoiding color collisions between rendered prompts and objects in the scene. However, this study did not verify whether these properties improve task performance. Real-world comparisons using scenes with red distractors and tasks that depend on background context are needed in future work to test these hypotheses.

\subsubsection{Human-Centered Evaluation and Multi-Turn Interaction}

This study did not directly evaluate whether the deictic interface reduces the user's descriptive burden or cognitive load. Validating this human-centered motivation requires a user study comparing task completion time, instruction length, number of corrections, and subjective workload across LI, VLI, and VI. Furthermore, multi-turn interaction, in which users can confirm the robot's interpretation of the target and placement goal and add or revise language or deictic gestures as needed, provides a direction for extending DeicticVLA to human--robot interaction.

In such an interactive configuration, an upstream VLM could propose candidate manipulation regions based on the user's language instruction. The user could then confirm or revise these regions before passing them to DeicticVLA, thereby dividing target identification and action generation between the two models. The upstream VLM would assist in determining what to manipulate, whereas the VLA would determine how to act. Another possible configuration would use LI when the upstream system can identify the target from language alone and VLI or VI when additional target specification by the user is required.


\section{Conclusion}
We proposed DeicticVLA, a unified interface that canonicalizes LI, VLI, and VI into a common representation consisting of a text prompt and deictic masks, enabling a single pretrained VLA to handle all three instruction modes. Controlled simulation comparisons showed that no single prompting method dominated across all evaluations, and that the prompting method should be selected according to the required generalization and operating conditions. The training-strategy ablations showed that two-stage training is important for acquiring the ability to use deictic masks in unseen layouts, and that retaining LI data in the second stage mitigates forgetting of language instruction following.

In the real-world experiments, a single policy supported all three instruction modes, and VLI and VI substantially outperformed LI under unseen instruction expressions, appearance changes, and novel objects, most notably achieving 100\% success for unseen categories where jointly trained LI achieved 16.7\%. The main remaining challenge is that jointly trained LI underperformed the LI-only baseline, indicating that the data balance among instruction modes must be improved for small-scale real-world datasets. Overall, DeicticVLA provides an extensible VLA interface in which complementary instruction modes can be used according to the task and environment.

\appendices
\section{Exploratory Evaluation of MP-Inter-$k$}
\label{app:mp_inter}

Beyond MP-Early and MP-Late, we conduct an exploratory evaluation of MP-Inter-$k$, which injects mask prompt embeddings after multi-head self-attention and before the feed-forward network at intermediate visual-encoder layers $\ell\in\mathcal K$:
\begin{equation}
\begin{aligned}
\mathbf Z_{\ell}'
&=
\mathbf Z_{\ell}
+\operatorname{MHSA}\!\left(
\operatorname{LN}\!\left(\mathbf Z_{\ell}\right)
\right),
\\
\mathbf Z_{\ell+1}
&=
\mathbf Z_{\ell}'
+\mathbb I[\ell\in\mathcal K]\mathbf Z_M
+\operatorname{FFN}\!\left(
\operatorname{LN}\!\left(
\mathbf Z_{\ell}'
+\mathbb I[\ell\in\mathcal K]\mathbf Z_M
\right)
\right).
\end{aligned}
\end{equation}

\noindent Here, $\operatorname{MHSA}$ and $\operatorname{FFN}$ denote multi-head self-attention and the feed-forward network, respectively, and $\mathbb I[\cdot]$ is the indicator function. For MP-Inter-3, we set $\mathcal K=\{0,9,18\}$ in the 27-layer visual encoder. Because $\mathbf Z_M=\mathbf 0$ for LI, these injections do not alter the LI image embeddings. We evaluate MP-Inter-3 under 2S, 2S-NoLI, and 1S; Table~\ref{tab:mp_inter_results} summarizes the results.

\begin{table}[h]
  \centering
  \caption{In-distribution results for MP-Inter-3.}
  \label{tab:mp_inter_results}
  \begin{tabular}{l rrr r rr}
    \toprule
    & \multicolumn{4}{c}{SR} & \multicolumn{2}{c}{$\Delta\mathrm{SR}$} \\
    \cmidrule(lr){2-5} \cmidrule(lr){6-7}
    Training & LI & VLI & VI & Mean & VLI & VI \\
    \midrule
    2S        & 62.0 & 41.4 & 22.0 & 41.8 & +1.6   & $-$2.7 \\
    2S-NoLI   & 49.9 & 34.6 & 25.9 & 36.8 & $-$2.7 & $-$7.1 \\
    1S        & 69.5 & 43.2 & 25.9 & 46.2 & $-$1.5 & $-$5.2 \\
    \bottomrule
  \end{tabular}
\end{table}

Across the three training strategies, MP-Inter-3 achieves substantially lower mean SR than the four methods in the main text and shows little or even negative contribution from deictic masks, while its final training loss is three to seven times as high. These results suggest that repeated intermediate-layer injection hindered optimization in this setting. Because we evaluate only $\mathcal K=\{0,9,18\}$, this failure does not rule out intermediate-layer fusion in general. We therefore exclude MP-Inter-3 from the main comparison and report it here as an exploratory failure case.

\bibliographystyle{IEEEtran}
\bibliography{references}

\end{document}